\documentclass[runningheads]{llncs}
\usepackage[T1]{fontenc}
\usepackage{graphicx}
\usepackage{amsmath} 
\usepackage{mathtools} 
\usepackage{bm}
\usepackage{cite}
\usepackage{amsmath,amssymb,amsfonts}
\usepackage{booktabs}
\usepackage{multirow}

\begin{document}

\title{Weakly Supervised Instance-Level \\ Gleason Pattern Estimation \\ Using Primary and Secondary Labels\\
}
\titlerunning{Weakly Supervised Instance-Level Gleason Pattern Estimation}

\author{Nao Sugeta\inst{1}\thanks{Equal contribution.} \and
Kaito Shiku\inst{1 *} \and
Shinnosuke Matsuo\inst{1} \and
Ryoma Bise\inst{1}}
\authorrunning{N. Sugeta et al.}

\institute{Kyushu University, Japan \\
    \email{nao.sugeta@human.ait.kyushu-u.ac.jp}\\
    \email{kaito.shiku@human.ait.kyushu-u.ac.jp}}

%
%


\maketitle

\begin{abstract}
In prostate cancer histopathology, the Gleason Score is determined by the most frequent (Primary) and second most frequent (Secondary) Gleason patterns within a whole-slide image. Although these slide-level labels are routinely available in clinical practice, instance-level Gleason annotations are rarely provided, making patch-level learning challenging. We propose a Multiple Instance Learning (MIL) framework that estimates instance-level Gleason patterns from slide-level Primary and Secondary labels. The proposed method formulates instance-level learning according to the clinical definition of the Gleason Score by aggregating instance predictions into class counts and explicitly modeling the Primary pattern, Secondary pattern, and their dominance. Experimental results demonstrate that the proposed formulation enables effective instance-level learning and outperforms existing MIL approaches on the SICAP-MIL dataset.

\end{abstract}

\section{Introduction}

\begin{figure}[t]
    \begin{center}
        \includegraphics[width=1\linewidth]{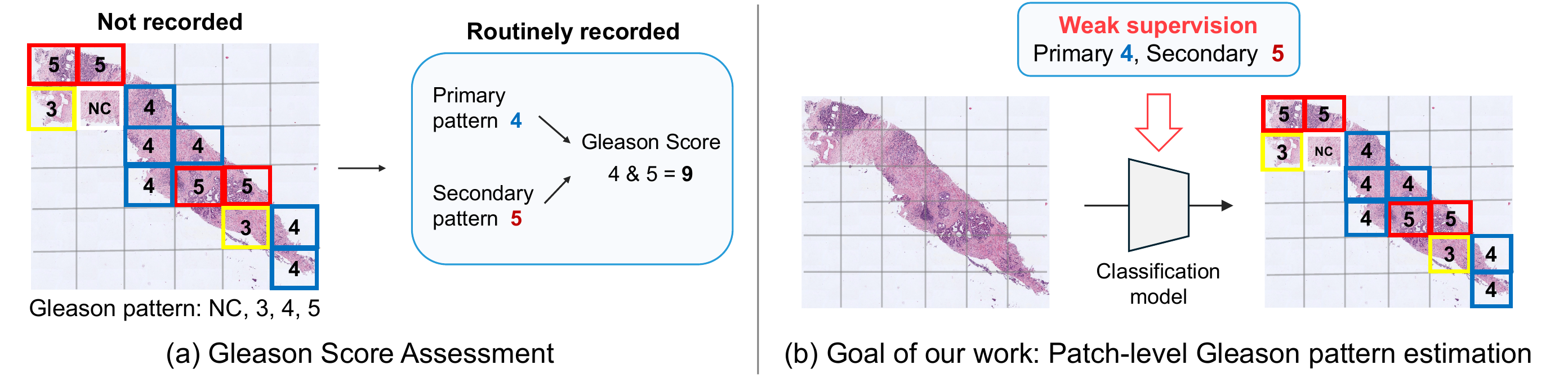}
        \vspace{-3mm}
        \caption{Overview of our work.}
        \vspace{-5mm}
        \label{fig:gleason}
    \end{center}
\end{figure}

Prostate cancer is one of the most prevalent malignancies worldwide~\cite{ferlay2013cancer}, and its clinical management critically depends on accurate assessment of tumor aggressiveness.
In routine pathology, this assessment is performed using the Gleason grading system~\cite{epstein20162014, bulten2022artificial}.
As illustrated in Fig.~\ref{fig:gleason} (a), each whole-slide image (WSI) is assigned two labels: the \emph{Primary} pattern, corresponding to the most frequent Gleason pattern, and the \emph{Secondary} pattern, corresponding to the second most frequent one.
The Gleason Score is defined as the sum of these two patterns and serves as a standard clinical indicator.

From a machine learning perspective, histopathological image analysis requires fine-grained, patch-level classification or segmentation.
As shown in Fig.~\ref{fig:gleason} (a), a WSI typically contains multiple regions exhibiting different Gleason patterns.
However, instance-level annotations are not usually available in clinical practice, and only the Primary and Secondary patterns are provided as slide-level supervision.
This mismatch between desired prediction granularity and available supervision poses a fundamental challenge.

Several Multiple Instance Learning (MIL) approaches have been proposed to leverage slide-level Gleason information~\cite{hao2025dual,anklin2021learning,bian2022multiple}.
While some methods directly predict Primary and Secondary patterns at the bag level, they do not infer labels for individual patches and therefore provide limited spatial interpretability.
In contrast, estimating instance-level Gleason patterns under weak supervision remains an open and clinically relevant problem.

In this work, as shown in Fig.~\ref{fig:gleason} (b), we propose a weakly supervised MIL framework that enables \emph{patch-level Gleason pattern estimation} using only slide-level Primary and Secondary labels.
To explicitly model the clinical definition of the Gleason Score, our method aggregates instance-level predictions by counting the predicted classes and optimizes them so that the resulting Primary and Secondary patterns are consistent with the ground-truth bag-level labels. Consequently, the model can learn local Gleason pattern assignments without explicit patch-level annotations.

The main contributions of this paper are summarized as follows:
\begin{itemize}
  \item We propose an instance-level MIL method that explicitly models the clinical definition of the Gleason Score based on the Primary pattern, Secondary pattern, and their dominance.
  \item The proposed approach produces spatially interpretable predictions while remaining consistent with the clinical definition of the Gleason Score.
  \item We demonstrate the effectiveness of the proposed method on the SICAP-MIL dataset~\cite{silva2022proportion} through quantitative and qualitative evaluations.
\end{itemize}

\section{Related Work}

Numerous studies have investigated Gleason Score prediction using machine learning for prostate cancer histopathology~\cite{arvaniti2018automated,nagpal2020development,hao2025dual,anklin2021learning,bian2022multiple,kaito2025learning}.
Early approaches relied on fully supervised learning with region-level annotations provided by expert pathologists~\cite{arvaniti2018automated,nagpal2020development}, which achieved high accuracy but required substantial annotation effort.

To reduce annotation cost, several studies have focused on learning from slide-level Gleason information, where the goal is to predict Gleason-related labels at the whole-slide level~\cite{hao2025dual,anklin2021learning,bian2022multiple}.
These methods typically adopt a MIL framework~\cite{kaito2025learning,wang2018revisiting,ilse2018attention,shao2021transmil}, treating each WSI as a bag of patches and directly predicting slide-level Gleason-related labels.

While effective for slide-level grading, these approaches are not designed to infer instance-level Gleason patterns and therefore provide limited spatial interpretability.
In parallel, instance-level MIL formulations have been explored in other contexts~\cite{wang2018revisiting,early2024inherently,kaito2025learning}, where instance predictions are aggregated into bag-level statistics using max or mean pooling~\cite{wang2018revisiting}, attention-based aggregation~\cite{early2024inherently}, or count-based aggregation~\cite{kaito2025learning}.
However, existing formulations typically focus on binary classification or a single dominant class and therefore do not reflect the clinical definition of the Gleason Score, which depends on the relative dominance of both the most frequent and the second most frequent Gleason patterns.
In contrast, our method enables instance-level Gleason pattern estimation under weak supervision while explicitly modeling both the Primary and Secondary dominance structure.

\section{Problem Formulation}
We study a prostate cancer histopathology classification problem under the Multiple Instance Learning (MIL) framework.
Each whole-slide image (WSI) is represented as a \emph{bag} consisting of multiple local image patches, referred to as \emph{instances}.

Each bag is associated with two bag-level labels derived from the Gleason grading system: the \emph{Primary} pattern, defined as the most frequent Gleason pattern within the bag, and the \emph{Secondary} pattern, defined as the second most frequent one.
According to the clinical definition of the Gleason Score, if a single Gleason pattern accounts for more than 95\% of the tissue, the Primary and Secondary labels coincide.
No class labels are provided for individual instances.

The objective is to learn an instance-level classifier using only these bag-level Primary and Secondary labels.
In this study, we focus on prostate cancer severity classification corresponding to Gleason patterns 3, 4, and 5.
Bags consisting entirely of non-cancerous tissue are excluded during training.

Formally, the training dataset is defined as
$\mathcal{D} = \{ \mathcal{B}^i, \bm{Y}_1^i, \bm{Y}_2^i \}_{i=1}^n$,
where $\mathcal{B}^i = \{ \bm{x}_j^i \}_{j=1}^{|\mathcal{B}^i|}$ denotes the $i$-th bag composed of instances $\bm{x}_j^i$.
The vectors $\bm{Y}_1^i, \bm{Y}_2^i \in \{0,1\}^C$ represent the one-hot encoded Primary and Secondary class labels of bag $i$, respectively, where $C$ is the number of classes.


\section{Proposed Method}
\begin{figure}[t]
    \begin{center}
        \includegraphics[width=1.\linewidth]{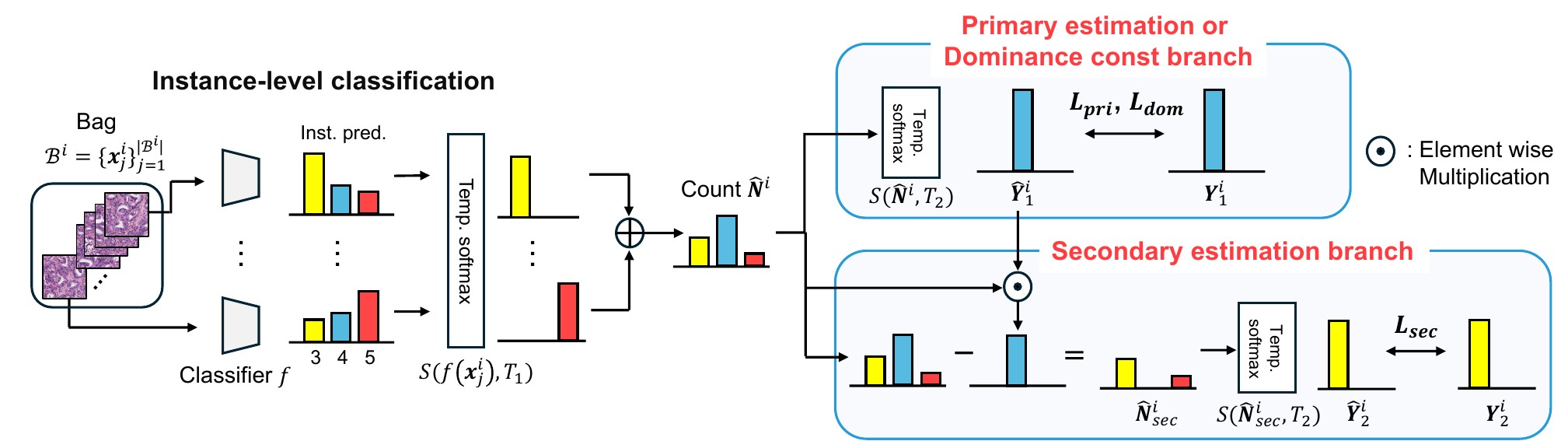}
        \caption{Overview of the Proposed Method.}
        \vspace{-4mm}
        \label{fig:proposed}
    \end{center}
\end{figure}

Fig.~\ref{fig:proposed} illustrates an overview of the proposed method.
The Gleason grading system defines the Primary and Secondary patterns based on the relative frequency of local Gleason patterns within a WSI.
To enable instance-level learning under such bag-level supervision, we build on dominance-based learning formulations that aggregate instance-level predictions to model a single dominant pattern~\cite{kaito2025learning}.
In contrast, the Gleason grading system requires modeling not only the most frequent pattern but also the second most frequent one.
Based on this observation, we propose a weakly supervised learning framework that explicitly accommodates the two-label structure of Gleason grading by separately estimating and supervising the Primary and Secondary patterns.

\subsection{Primary Pattern Estimation and Loss}

We first describe how the Primary Gleason pattern is estimated.
Given a bag corresponding to a WSI, each instance (patch) is independently fed into an instance-level classifier to predict a class assignment.
These instance-level predictions are converted into near one-hot class vectors and aggregated across the bag to count the occurrences of each Gleason pattern.
The Primary pattern is then determined as the class with the largest count.
To make this counting process differentiable and suitable for end-to-end training, we employ a temperature-scaled softmax function to approximate hard class assignments.
Since instance-level annotations are not available, the instance-level classifier is trained solely through gradients propagated from bag-level Primary and Secondary losses.

Let $g : \mathbb{R}^{w \times h \times d} \rightarrow [0,1]^C$ denote an instance-level classifier, where $C$ is the number of classes.
The classifier consists of a neural network
$f : \mathbb{R}^{w \times h \times d} \rightarrow \mathbb{R}^C$
followed by a temperature-scaled softmax function
$S : \mathbb{R}^C \rightarrow [0,1]^C$.
Given an instance $\bm{x}_j^i$, the classifier outputs
\begin{equation}
\bm{a}_j^i = g(\bm{x}_j^i) = S(f(\bm{x}_j^i), T_1),
\end{equation}
where $T_1$ is set to a sufficiently low temperature so that $\bm{a}_j^i$ approximates a one-hot vector.

For a bag $\mathcal{B}^i = \{ \bm{x}_j^i \}_{j=1}^{|\mathcal{B}^i|}$, class-wise prediction counts are obtained by aggregating instance-level outputs:
\begin{equation}
N_c^i = \sum_{j=1}^{|\mathcal{B}^i|} a_{j,c}^i ,
\end{equation}
where $a_{j,c}^i$ denotes the $c$-th element of $\bm{a}_j^i$.
The resulting count vector is denoted as
$\bm{N}^i = (N_1^i, \ldots, N_C^i)^{\mathsf{T}}$.

Based on the count distribution $\bm{N}^i$, the Primary pattern is defined as the class with the largest count.
To enable gradient-based optimization, the Primary label is estimated in a differentiable manner using a temperature-scaled softmax:
\begin{equation}
\hat{\bm{Y}}_1^i = S(\bm{N}^i, T_2),
\end{equation}
where $\hat{\bm{Y}}_1^i$ approximates a one-hot vector indicating the primary Gleason pattern, and $T_2$ is a sufficiently low temperature.

The Primary pattern is supervised using a bag-level loss so that the most frequent predicted pattern matches the ground-truth Primary label.
Specifically, we apply the standard cross-entropy loss to the estimated Primary label:
\begin{equation}
\mathcal{L}_{\mathrm{pri}}
= \mathcal{L}_{\mathrm{CE}}(\hat{\bm{Y}}_1^i, \bm{Y}_1^i).
\end{equation}

\subsection{Secondary Pattern Estimation and Loss}

The Secondary pattern is defined as the second most frequent Gleason pattern within a bag.
To estimate it under weak supervision, we remove the contribution of the Primary pattern from the class count distribution and identify the next dominant pattern in a fully differentiable manner.

Let $\hat{\bm{N}}^i$ denote the differentiable estimate of the class count distribution for bag $\mathcal{B}^i$.
As in the Primary case, the Primary pattern is first approximated using a temperature-scaled softmax applied to the estimated count distribution, such that the element corresponding to the largest count takes a value close to 1.
This softmax output serves as a differentiable approximation of the \emph{estimated} Primary class indicator.
We then define a masked count distribution by suppressing the estimated Primary component as
\begin{equation}
\hat{\bm{N}}_{\mathrm{sec}}^i
= \hat{\bm{N}}^i - \hat{\bm{N}}^i \odot S(\hat{\bm{N}}^i, T_3),
\end{equation}
where $\odot$ denotes element-wise multiplication and $T_3$ is a sufficiently low temperature.
This operation effectively removes the contribution of the Primary pattern while preserving end-to-end differentiability.

Subsequently, the secondary pattern is obtained in the same manner as the primary pattern estimation by applying a temperature-scaled softmax to the masked distribution:
\begin{equation}
\hat{\bm{Y}}_2^i = S(\hat{\bm{N}}_{\mathrm{sec}}^i, T_2),
\end{equation}
where this operation yields $\hat{\bm{Y}}_2^i$ as an approximation of a one-hot vector indicating the secondary Gleason pattern.


The Secondary pattern is supervised using the ground-truth Secondary label $\bm{Y}_2^i$.
Accordingly, the Secondary loss is defined as
\begin{equation}
\mathcal{L}_{\mathrm{sec}}
= \mathcal{L}_{\mathrm{CE}}(\hat{\bm{Y}}_2^i, \bm{Y}_2^i).
\end{equation}

\subsection{Dominance Constraint for Single-Pattern Slides}
According to the clinical definition of the Gleason Score, when a single Gleason pattern occupies more than 95\% of a WSI, the Primary and Secondary labels coincide, i.e., $\bm{Y}_1^i = \bm{Y}_2^i$.
This dominance criterion is part of the standard diagnostic protocol rather than a heuristic design choice.
To explicitly incorporate this clinically defined condition into the training objective, we introduce a dominance constraint loss that is applied only to slides for which the ground-truth Primary and Secondary labels are identical.

We obtain the class proportion vector $\bm{p}^i$ by normalizing the class counts.
Let $c_p$ denote the index of the Primary class indicated by the one-hot vector $\bm{Y}_1^i$.
The dominance constraint loss is defined using an indicator function $\mathbb{I}(\cdot)$ as
\begin{equation}
\mathcal{L}_{\mathrm{dom}}
=
\mathbb{I}\!\left( \bm{Y}_1^i = \bm{Y}_2^i \right)
\mathbb{I}\!\left( p_{c_p}^i < 0.95 \right)
\, \mathcal{L}_{\mathrm{CE}}(\bm{p}^i, \bm{Y}_1^i),
\end{equation}
which penalizes cases where the dominance condition is violated despite the Primary and Secondary labels being identical.
When $\bm{Y}_1^i \neq \bm{Y}_2^i$ or $p_{c_p}^i \ge 0.95$, the constraint loss becomes zero, reflecting cases in which the clinical dominance rule is satisfied.

\subsection{Overall Training Objective}

The final training objective is defined as a weighted sum of the above loss terms:
\begin{equation}
\mathcal{L}
= \mathcal{L}_{\mathrm{pri}} + \mathcal{L}_{\mathrm{dom}} + \lambda \mathcal{L}_{\mathrm{sec}},
\end{equation}
where $\lambda$ controls the contribution of the Secondary loss.

\section{Experiments}
\subsection{Dataset and Experimental Setup}

We evaluate our method on SICAP-MIL~\cite{silva2022proportion}, a prostate cancer histopathology dataset containing 350 WSIs.
Each WSI is annotated with bag-level Gleason information (Primary and Secondary patterns), and a subset additionally provides instance-level labels for four classes: non-cancer (NC) and Gleason patterns 3, 4, and 5.

We use 252 WSIs for training, supervised exclusively by bag-level Primary and Secondary labels.
Instance-level annotations are not used for training.
For validation and evaluation, we use 9 WSIs with instance-level annotations as a validation set and 79 WSIs as a held-out test set.
An additional 10 WSIs with instance-level annotations are used only to fine-tune the CONCH model~\cite{ming2024visual} for cancer versus non-cancer classification, ensuring no data leakage.

During training, instances predicted as cancerous by CONCH are used to form bags for learning Gleason patterns 3, 4, and 5.
During evaluation, instance-level predictions are obtained in a two-stage manner: CONCH first predicts cancer versus non-cancer, and cancerous instances are further classified into Gleason patterns 3, 4, or 5 by the proposed model.
Performance is evaluated on all four classes (NC, 3, 4, and 5) at the instance level.

\begin{table}[t]
    \centering
    \caption{Comparison of instance-level Gleason pattern classification performance.}
    \vspace{-2mm}
    \begin{tabular}{c@{\hspace{8mm}}c@{\hspace{8mm}}c}
        \toprule
        Method & Kappa & Macro-F1 \\
        \midrule
        Max~\cite{wang2018revisiting} &
        $0.701 \pm 0.018$ &
        $0.627 \pm 0.014$ \\

        Mean~\cite{wang2018revisiting} &
        $0.738 \pm 0.007$ &
        $0.678 \pm 0.009$ \\

        MILLET~\cite{early2024inherently} &
        $0.737 \pm 0.006$ &
        $0.678 \pm 0.008$ \\

        LML~\cite{kaito2025learning} &
        $0.727 \pm 0.040$ &
        $0.678 \pm 0.043$ \\

        Ours &
        $\mathbf{0.743} \pm 0.005$ &
        $\mathbf{0.697} \pm 0.009$ \\
        \bottomrule
    \end{tabular}
    \label{tab:instance_acc}
\end{table}

\begin{table}[t]
    \centering
    \caption{Ablation study of the proposed loss components.}
    \vspace{-2mm}
    \begin{tabular}{c@{\hspace{5mm}}
                    c@{\hspace{5mm}}
                    c@{\hspace{5mm}}
                    c@{\hspace{8mm}}
                    c@{\hspace{8mm}}
                    c}
        \toprule
         &
        $L_{\mathrm{pri}}$ &
        $L_{\mathrm{dom}}$ &
        $L_{\mathrm{sec}}$ &
        Kappa &
        Macro-F1 \\
        \midrule

        \multirow{3}{*}{Ours}
        & $\checkmark$ & & &
        $0.727 \pm 0.040$ &
        $0.678 \pm 0.043$ \\

        & $\checkmark$ &
        $\checkmark$ &
        &
        $0.739 \pm 0.007$ &
        $0.691 \pm 0.009$ \\

        & $\checkmark$ &
        $\checkmark$ &
        $\checkmark$ &
        $\mathbf{0.743} \pm 0.005$ &
        $\mathbf{0.697} \pm 0.009$ \\

        \bottomrule
    \end{tabular}
    \label{tab:instance_acc_ablation}
\end{table}

\subsection{Implementation Details}

All experiments are implemented using PyTorch~\cite{Paszke2019PyTorchAI}, with CONCH used as the feature extractor for the classifier $f$.
The Adam optimizer is used with a learning rate of $3 \times 10^{-6}$, a batch size of 16, and training for 10,000 epochs.
The temperature parameters are set to $T_1, T_2 = 0.1$ and $T_3 = 0.01$, and the Secondary loss weight $\lambda$ is set to 0.1.
Model selection is based on the best validation Macro-F1 score, and results are averaged over five random seeds.

\subsection{Experimental Results}

\noindent
{\bf Comparisons.}
To evaluate the effectiveness of the proposed method, we compared it with four MIL-based instance-level classification methods.
{\bf Max}, {\bf Mean} \cite{wang2018revisiting}, and {\bf MILLET}~\cite{early2024inherently}: Instance-level predictions are aggregated using max pooling, mean pooling, or attention, respectively, and trained with a bag-level multi-label loss.
{\bf LML}~\cite{kaito2025learning}: A count-based aggregation strategy trained using only the majority label of each bag.

Table~\ref{tab:instance_acc} presents the comparison results. Max pooling performs substantially worse than the other methods, likely because its aggregation strategy is not aligned with the diagnostic principles of Gleason grading. Although Mean pooling and MILLET achieve better performance, they do not explicitly model the relative proportions of primary and secondary Gleason patterns. LML captures the dominance of the primary pattern but fails to account for the secondary pattern, resulting in inferior performance.
In contrast, the proposed method achieves the best performance by explicitly incorporating the diagnostic principles of Gleason grading, including the relative roles of primary and secondary patterns as well as pattern dominance.


%

\noindent
{\bf Eﬀectiveness of proposed method.}
Table~\ref{tab:instance_acc_ablation} summarizes an ablation study on the proposed loss design, evaluating the contribution of each component to instance-level classification performance.
Using only the Primary loss $\mathcal{L}_{\mathrm{pri}}$ provides a baseline level of accuracy.
Introducing the dominance constraint $\mathcal{L}_{\mathrm{dom}}$, which enforces stronger consistency with the Primary pattern, leads to a clear improvement across all metrics.
The best performance is achieved when the Secondary loss $\mathcal{L}_{\mathrm{sec}}$ is further incorporated, indicating that explicitly modeling the Secondary pattern in addition to the Primary pattern enhances instance-level prediction.
These results demonstrate that jointly leveraging Primary supervision, the dominance constraint, and Secondary supervision enables effective exploitation of bag-level Gleason information for instance-level Gleason pattern prediction.

\begin{figure}[t]
    \begin{center}
        \includegraphics[width=0.78\linewidth]{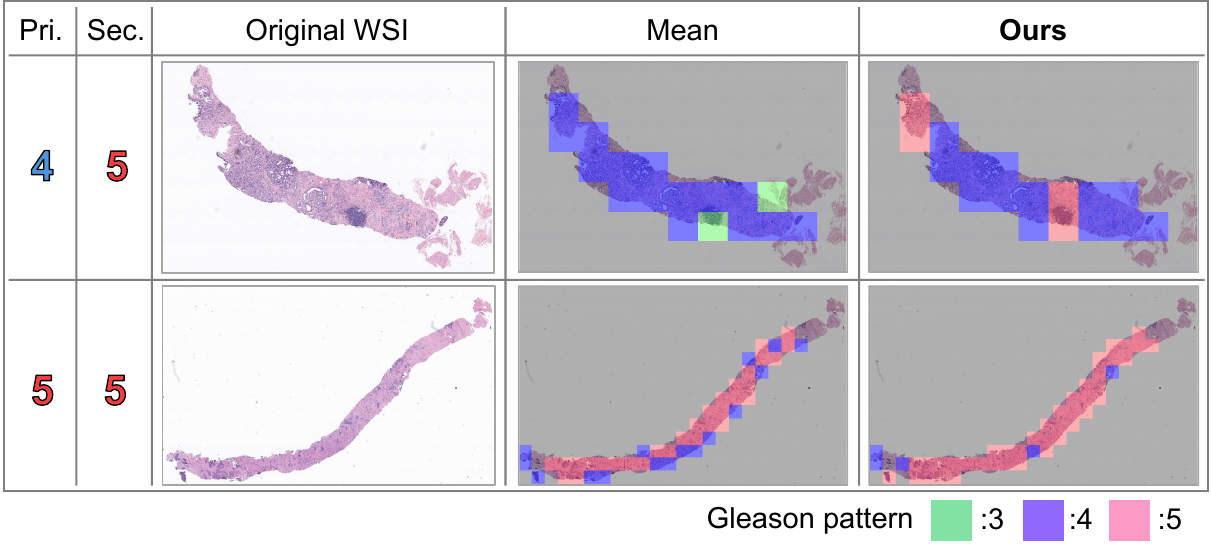}
    \vspace{-4mm}
        \caption{Qualitative instance-level Gleason pattern predictions visualized as segmentation maps. Samples in the top row correspond to primary pattern 4 and secondary pattern 5, while those in the bottom row are dominated by Gleason pattern 5 (>95$\%$). Gleason patterns 3, 4, and 5 are shown in green, blue, and red, respectively.}
        \label{fig:seg}
    \end{center}
    \vspace{-4mm}
\end{figure}

To qualitatively demonstrate that the proposed method produces predictions consistent with the clinical definition of the Gleason Score, Fig.~\ref{fig:seg} visualizes the segmentation results generated by the proposed method and a conventional MIL approach based on mean pooling.
Gleason patterns 3, 4, and 5 are shown in green, blue, and red, respectively.
As shown in the top row of Fig.~\ref{fig:seg}, the mean-pooling-based method fails to correctly identify the secondary Gleason pattern in a sample with primary pattern 4 and secondary pattern 5. In contrast, the proposed method successfully captures both patterns, producing predictions consistent with the Gleason grading criteria.
Furthermore, as shown in the bottom row of Fig.~\ref{fig:seg}, for a sample dominated by Gleason pattern 5 (>95$\%$), the mean-pooling-based method predicts a substantial proportion of regions as other patterns, resulting in a distribution inconsistent with the underlying pathology. In contrast, the proposed method, aided by the dominance constraint, more accurately captures the dominant presence of Gleason pattern 5.

\section{Conclusion}
We proposed a weakly supervised MIL framework for instance-level Gleason pattern estimation using only clinically available Primary and Secondary Gleason labels. Rather than directly predicting slide-level labels, the proposed method formulates instance-level learning according to the clinical definition of the Gleason Score by explicitly modeling the Primary pattern, Secondary pattern, and their dominance within a count-based MIL framework.
Experiments on the SICAP-MIL dataset demonstrated that the proposed formulation improves instance-level classification performance while producing spatially interpretable predictions.



\noindent
{\bf Prospect of application:} 
The proposed framework can support computer-aided prostate cancer diagnosis by estimating patch-level Gleason patterns using only routinely available slide-level Primary and Secondary labels. It has the potential to reduce annotation costs while improving spatial interpretability in digital pathology workflows and large-scale retrospective studies.


\section{Disclosure of Interests}
The authors have no competing interests to declare.

\section{Acknowledgments}
This work was supported by  JST BOOST Grant Number JPMJBS2406,  JST ACT-X Grant Number JPMJAX23CR, ASPIRE Grant Number JPMJAP2403 and AMED 26mk0121326h0X02.


\bibliographystyle{splncs04}
\bibliography{myrefs}


\end{document}